\documentclass[a4paper, 10pt, conference]{ieeeconf}      
\usepackage{FG2020}

\FGfinalcopy 

\usepackage{cite}
\usepackage{amsmath,amssymb,amsfonts}
\usepackage{algorithmic}
\usepackage{textcomp}

\usepackage{soul} 
\usepackage{tcolorbox}
\usepackage{graphicx}
\usepackage{import}
\usepackage{xcolor}
\usepackage{float} 

\IEEEoverridecommandlockouts                              
\def\FGPaperID{17} 

\title{\LARGE \bf
Affective Expression Analysis in-the-wild using Multi-Task Temporal Statistical Deep Learning Model
}

\author{\parbox{16cm}{\centering
    {\large Nhu-Tai Do, Tram-Tran Nguyen-Quynh and Soo-Hyung Kim$^*$}\\
    {\normalsize
    School of Electronics and Computer Engineering, Chonnam National University\\
    77 Yongbong-ro, Buk-gu, Gwangju 500 – 757, Korea\\
    donhutai@gmail.com, shkim@jnu.ac.kr}}
    \thanks{$^*$The corresponding author is Soo-Hyung Kim.}
}

\begin{document}

\ifFGfinal
\thispagestyle{empty}
\pagestyle{empty}
\else
\author{Anonymous FG2020 submission\\ Paper ID \FGPaperID \\}
\pagestyle{plain}
\fi
\maketitle

\begin{abstract}
Affective behavior analysis plays an important role in human-computer interaction, customer marketing, health monitoring. ABAW Challenge and Aff-Wild2 dataset raise the new challenge for classifying basic emotions and regression valence-arousal value under in-the-wild environments. In this paper, we present an affective expression analysis model that deals with the above challenges. Our approach includes STAT and Temporal Module for fine-tuning again face feature model. We experimented on Aff-Wild2 dataset, a large-scale dataset for ABAW Challenge with the annotations for both the categorical and valence-arousal emotion. We achieved the expression score 0.543 and valence-arousal score 0.534 on the validation set.

\end{abstract}

\section{INTRODUCTION}
Understanding affective behavior analysis is an active research due to its fundamental role in wide applications such as  human-computer interaction, customer marketing, health monitoring, etc. Through the many years, it is a challenging task due to complex and dynamic properties in expression as well as diverse environments in-the-wild.  

The important target of the affective behavior analysis focuses on helping the machine to be able to understand the human emotions or emotion (expression) recognition. The most popular emotion representation is to describe by seven basic emotions such as Neutral, Angry, Disgust, Fear, Happy, Sad, Surprise by Paul Ekman's work \cite{Ekman1992b}. The differences from the emotions are based on the properties of distinctive  physiology, universal  signals, thoughts, memories, images, etc. The other representation of human emotion is described in continuous space using the 2D Valence-Arousal Emotion Wheel. The valence axis measures the level of pleasure. Besides, the arousal axis indicates the level of affective activation \cite{Plutchik1984}.

Recently, to promote the development of the Affective Behavior Analysis problems and the requirements of the huge real-world data for deep learning approach, Kollias et al. provide the large-scale dataset Aff-Wild2 \cite{Kollias2018c, zafeiriou2017aff}, propose many baseline methods \cite{kollias2019deep, Kollias2019a, Kollias2018b, kollias2017recognition}, and organize the Affective Behavior Analysis in-the-wild (ABAW) \cite{Kollias2020}.  

The Aff-Wild2 \cite{Kollias2018c} is the extension version of Aff-Wild \cite{zafeiriou2017aff} to the large-scale dataset. It is collected the huge in-the-wild videos from  YouTube with the wide-range subjects about age, ethnicity, profession, head pose, illumination conditions, etc. Moreover, it contains the annotations with valence and arousal, discrete emotions and action units. 

In this paper, we describe the proposed method for joining track 1 about valence-arousal regression and track 2 about categorical emotion recognition in the ABAW challenge. We analyze data and recognize it contains some difficulties. 

\begin{figure}[t]
	\center{\includegraphics[width=0.6\linewidth]{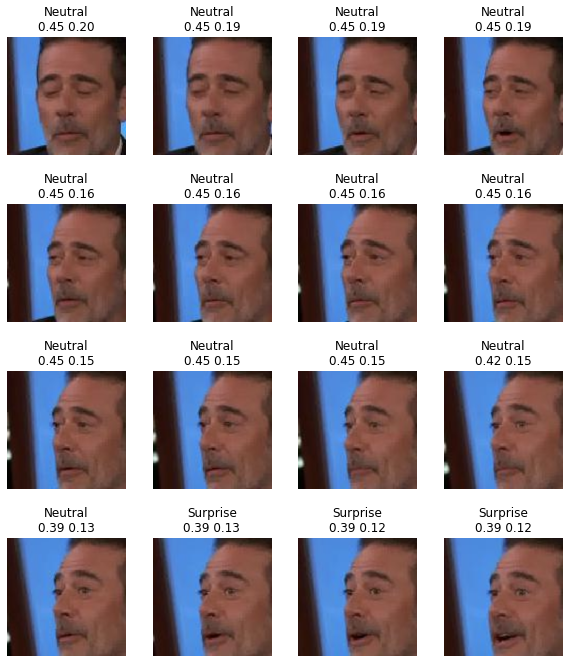}}
	\caption{Affective Behaviour Recognition difficulties. The bottom-right and bottom-left images are nearly the same about arousal-valence values but different from the categorical emotion with Surprise and Neutral. Besides, the top-left and bottom-left images are the same categorical emotion but big different from arousal-valence values.}
	\label{fig:problem}
\end{figure}

\begin{figure*}
	\center{\includegraphics[width=0.85\linewidth]{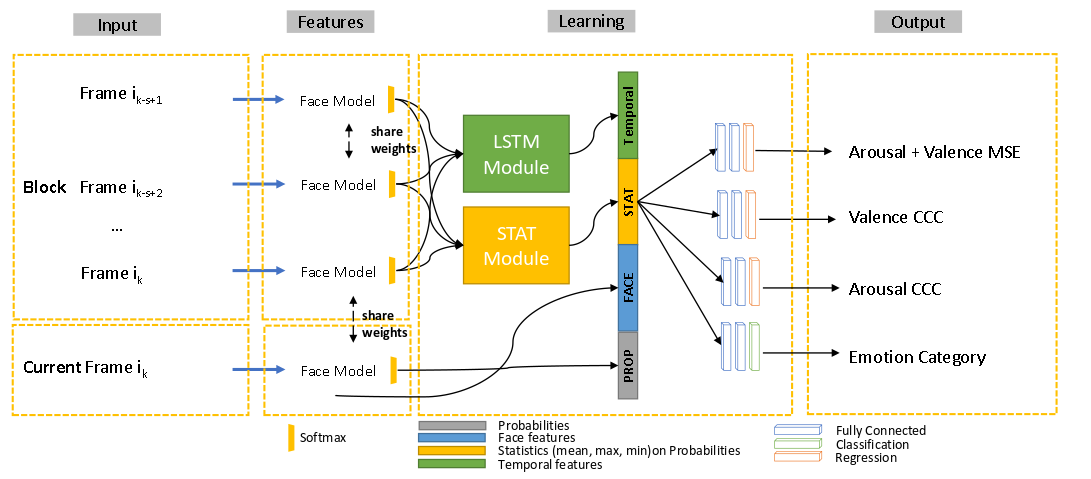}}
	\caption{Overview of the proposed system for prediction the categorical emotion and arousal-valence regression. The STAT module will merge the prediction probability scores from the previous continuous frames by mean, max, and min operator. LSTM module consists one or two LSTM cell to exploit the temporal relation of the probability scores from the frame blocks. After that, all features from STAT, LSTM module and probability score, face features of current frames will be merged. The classification and regression blocks will receive the fusion features to output results.}
	\label{fig:proposed_model}
\end{figure*}

Next, the task predicts the emotion by frame-by-frame with small displacements in the faces as Fig.\ref{fig:problem}. It is different from the AFEW dataset \cite{dhall2012collecting} and EmotiW Challenge \cite{Dhall2019b} with the ground-truth for the video emotion. So, it is difficult to recognize the emotion between two consecutive frames in the same video. In Fig.\ref{fig:problem}, we see the bottom-right image is nearly the value of valence-arousal emotion with the bottom-right image. But it is different from the category emotion with surprise emotion (for the bottom-right image) against neutral emotion (for the bottom-left image). Moreover, the top-left and bottom-left images are the same categorical emotion but they are the big difference from valence-arousal emotion. Finally, the problem needs to deal with the imbalance dataset on neutral emotion (very high) and angry, disgust emotion (very small).

To overcome the in-the-wild environment, we use the pre-trained weight on the VGG-Face2 Resnet50 model for training on AffectNet and RAF-DB dataset. We transfer learning the model learned from the well-known and good datasets to the Aff-Wild2 dataset. 

To deal with the unbalanced-data problem, we use the balance emotion sampling and online data augmentation technique on every batch during the training process. 

To enhance under the noises in the ground-truth among consecutive frames, we use the statistical encoding to merge the features of the previous consecutive frames by the min, max, average. After that, we concatenate all the features from the current frame, the temporal features from the previous consecutive frames, and the statistical encoding features for multi-task learning. Multi-task learning contains categorical emotion classification and valence-arousal emotion regression. It has the role of the regularizing effect based on shared features among classification and regression tasks.

The paper is organized as follows. In Section 2, we provide the details of our proposed method for track 1 and 2 of the challenge. Then we show experimental results and discussion. Finally, we conclude our research and discuss further works.

\section{PROPOSED METHOD}


\subsection{Problem Overview}
Given an input face frame $F_{i,k} \in R^{h \times w \times 3}$ at time $k$ of a video sequence $v_{i}$, our objective is to predict the face frame to the valence-arousal emotion value  $c_{i,k} \in \left[-1, 1\right]$ and categorical emotion $e_{i, k} \in \left[0, 6\right]$ corresponding to $\left\{ Neutral, Angry, Disgust, Fear, Happy, Sad, Surprise \right\}$. This problem needs not only to detect continuous and discrete emotion in the current frame with the spatial dimension but also to exploit the similarities among consecutive frames in the same video.  

To tackle this problem, we first train the VGGFace2 network on AffectNet and RAF-DB datasets for transfer learning. This model has the role as the feature extraction with two outputs: face emotion feature representation and emotion probability scores. 

Besides, we assume the emotion at the current frame only affected by previous frames. The reason is that during the ground-truth process, the labelers make a decision at the current frame only affected by the view of the previous frames in the same video. This assumption also helps the model to reduce the input space and improve performance. 

Therefore, the input of the proposed model as Fig.\ref{fig:proposed_model} is the current frame $F_{i, k}$ and the previous consecutive frames $F_{i, k}, F_{i, k-1}, ..., F_{i, k - s + 1}$ where $s$ is the number of selection frames. We use probability scores from the face feature model for exploiting the temporal and statistical relation by two module LSTM and STAT. The features from two modules are merged with the features of the current frame for predicting the result.

Finally, we use the multi-task loss for the regularizing effect between the categorical emotion values and the valiance arousal continuous emotion values. 

\subsection{Face Feature Model}
In our proposed method, we utilize ResNet50 network \cite{He2016} pre-trained on VGGFace2 \cite{cao2018vggface2} as a feature extraction network. The ResNet50 network is a  conventional convolutional neural network that trained on the million face images from the Internet of the large-scale VGGFace2 dataset. It has 50 layers deep using 4 stages of the convolutional and identity blocks to classify the face image into 8631 classes for person recognition. 

We modify the pre-trained VGGFace2 network by eliminating the last layer and inserting the new classification layer with 7 classes for basic categorical emotion. From there, we fine-tune again on AffectNet and RAF-DB datasets.  As a result, the network can learn rich feature representations from the large-scale face emotion datasets.

\subsection{Proposed Model}
Our proposed model as Fig.\ref{fig:proposed_model} consists of the face feature model, LSTM and STAT Module, feature fusion section and classification \& regression section. 

For the face feature model, we use the pre-trained VGGFace2 ResNet50 network described in the previous section. We freeze all weight values exception the last stage convolution and identity block as well as the classification layer. The current frame and the block of previous consecutive frames are inputted to the model for outputting the feature and probability scores.

The LSTM Module will receive the probability scores of the previous frames to exploit the temporal relationship among the frames. It consists of one or two bidirectional LSTM cells with 1024 length, and return the temporal features. 

Similarly, the STAT module will take the probability scores and face features and calculate the mean, max, average from input values. From there, the module gives the statically attributes of the previous frames. It will help our model to prevent the noise and learn the statistical features from the group of frames.

All temporal and statistical features from the group of frames will be fused with the probability scores and face features of the current frame. Fusion features will input to one classification branch and three regression branches. The classification branch has two dense layers and one soft-max layer for outputting seven classes of basic emotion. The three regression branches have two dense layers and the last dense layer using tanh activation. The regression outputs include arousal, valence value for Concordance Correlation Coefficient (CCC) loss, one vector with length 5 corresponding to arousal, valence, one average value, and two different values for mean square error (MSE) loss.

\subsection{Multi-task Loss}
For the basic emotion classification, we use the categorical cross-entropy loss as follows:
\begin{equation}
    \pounds_{class}=-\sum_{i=1}^{C}y_{i}log\hat{y}_{i}    
    \label{eq:categorical_cross_entropy_loss}
\end{equation}
where $y_{i}$ is the one-hot vector of the ground-truth of the basic emotion, $\hat{y}_{i}$ is the predicted probability vector, and C is the seven emotion.

For the arousal and valence regression, we use the Concordance Correlation Coefficient loss as follows:
\begin{equation}
    \pounds_{ccc} = 1 - \frac{2s_{y\hat{y}}}{\sigma_{y}^{2} + \sigma_{\hat{y}}^{2} + \left(\bar{y}-\bar{\hat{y}}\right)^{2}}
    \label{eq:concordance_correlation_coefficient_loss}
\end{equation}
where $y$ is the ground-truth values, $\hat{y}$ is the prediction values; $\bar{y}$, $\bar{\hat{y}}$ are the mean values of $y$, $\hat{y}$ respectively; $\sigma_{y}^{2}$, $\sigma_{\hat{y}}^{2}$ are the variance values of $y$, $\hat{y}$ respectively; and $s_{y\hat{y}}$ is the covariance of $y$, $\hat{y}$.

For the regularity effect to the arousal and valence regression, we build the vector of arousal, valence values, average of arousal and valence, and two different values between arousal and valence with its average value. After that, we use the mean square error loss for the estimation between the ground-truth and prediction on this vector as follow:
\begin{equation}
    \pounds_{mse}=\sum_{i=1}^{N}\left(y_{i} -\hat{y}_{i} \right )^{2}
    \label{eq:mean_square_error_loss}
\end{equation}
where $y_{i}$ is the ground-truth vector, and $\hat{y}_{i}$ is the predicted vector calculated from arousal and valence values.

Finally, our network combines $\pounds_{class}$, $\pounds_{arousal, ccc}$, $\pounds_{valence, ccc}$ and $\pounds_{mse}$ as follows:
\begin{equation}
    \pounds_{total} = w_{1}\pounds_{class} + w_{2}\pounds_{arousal, ccc} + w_{3}\pounds_{valence, ccc} + w_{4}\pounds_{mse}
    \label{eq:total_loss}
\end{equation}

In this paper, we set $w_{1} = 1.0$, $w_{2} = w_{3} = 0.4$, and $w_{4} = 0.2$.

\begin{table*}
\centering
\caption{List all derivation models from our proposed model for ablation studies}
\label{tbl:list_models}
\begin{tabular}{|c|l|c|c|}
\hline
\textbf{Model No.}   &  \multicolumn{1}{c|}{\textbf{Name}} & \multicolumn{1}{c|}{\textbf{Input}} & \textbf{Output} \\ \hline
1   & Emotion Image         & Image     & Expression \\ \hline
2   & Emotion VA Image      & Image     & Expression, Valence-Arousal \\ \hline
3   & Emotion Frame         & Image, previous blocks    & Expression \\ \hline
4   & Emotion VA Frame      & Image, previous blocks    & Expression, Valence-Arousal \\ \hline
5   & Emotion Frame with LSTM & Image, previous blocks  & Expression \\ \hline
6   & Emotion VA Frame with LSTM & Image, previous blocks    & Expression, Valence-Arousal \\ \hline
\end{tabular}
\end{table*}

\section{EXPERIMENTS AND DISCUSSION}
\subsection{Datasets and Environments}
For face feature model, we fine-tuned on Affect-Net and RAF-DB datasets. In the AffectNet dataset, the images are chosen with the only seven labels same as the Aff-Wild2 dataset. There are 283,901 images for training, and 3,500 images for validation. In the RAF-DB dataset, there are 12,271 for training and 3,068 for validation. 

Aff-Wild2 is the dataset used in ABAW Challenge. There are three tracks: Valence-Arousal Regression, Basic Emotion Recognition and Emotion Action Unit Recognition. In track basic emotion recognition, we eliminated these frames without the annotations in training and validation. So, there are 917835 images and 251 videos in the training set,  318503 images, and 69 videos in the validation set as well as 997332 images and 223 videos in the testing set. 

\begin{figure}[t]
	\center{\includegraphics[width=0.9\linewidth]{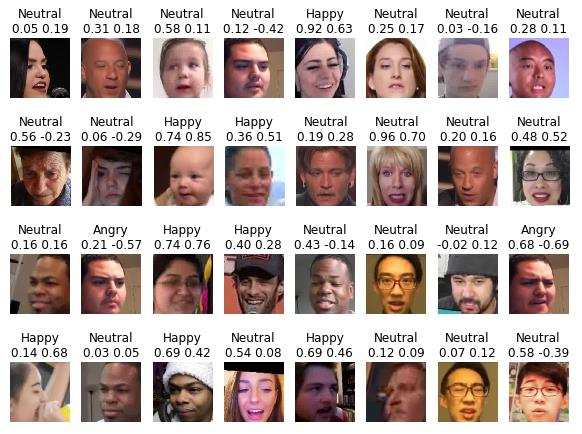}}
	\caption{The Aff-wild2 frame images. It shows a variety of age, pose, illumination, occlusion under in-the-wild environments. Furthermore, it is difficult to distinguish the emotions in some images such as the rightmost second-row image has neutral emotion. But it is nearly a happy-emotion image with the valence-arousal value close to the value of the fourth image in the third row.}
	\label{fig:affwild2_images}
\end{figure}

Fig. \ref{fig:affwild2_images} shows Aff-Wild2 has the face images under in-the-wild environments with a variety of age, pose, illumination, occlusion, etc. Especially, there are many neutral images are near the images in the other emotions such as the fourth image at the bottom-left in Fig. \ref{fig:affwild2_images} almost as happy emotion. It is the truth for the images in the same video as Fig.\ref{fig:problem}.

\begin{figure}[H]
	\center{\includegraphics[width=0.8\linewidth]{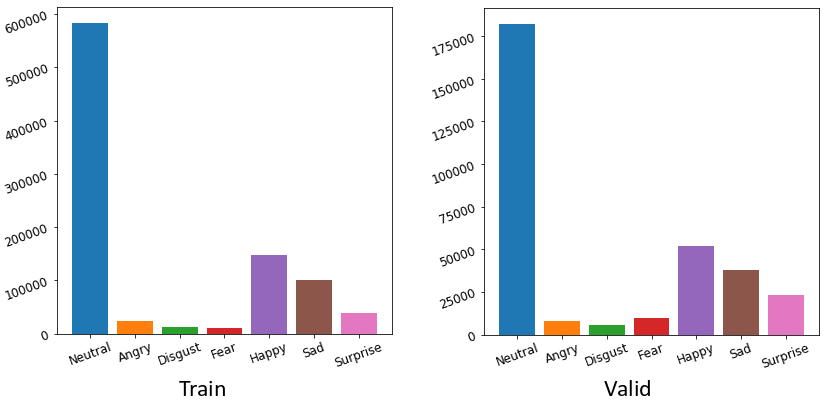}}
	\caption{Data distribution in the basic emotion recognition track of training and validation set.}
	\label{fig:affwild2_emotion_scheme_data_distributio}
\end{figure}

The data distribution in the basic emotion recognition track as Fig. \ref{fig:affwild2_emotion_scheme_data_distributio}. It shows the imbalanced data in the dataset. Neutral images have more than the other emotions. The Angry, Disgust and Fear have the less than Happy, Sad and Surprise emotion.

For the valence-arousal track, we built the training and validation set with full of the basic emotion label and valence-arousal label in the training and validation set of the basic emotion recognition track. About testing set, we kept two testing set in the basic emotion recognition and valence-arousal regression track. So, the training and validation set have 194 videos with 720365 images and 57 videos with 254674 images, respectively. For testing set, there are 223 videos with 997332 images in the basic emotion recognition and 139 videos with 714986 images in the valence-arousal regression track.

For the setup environment, we used Tensorflow Keras on Python 3.5 to build our proposed model. We chose SGD, Adam optimizer for training with learning rate 0.0001 for all models and took the best model after hyper-parameter tuning.

\subsection{Ablation Studies}
We built six derivation models from our proposed model listed in Table \ref{tbl:list_models}. The model 1 (Emotion Image), and 2 (Emotion VA Image) is frame-by-frame input. They only had the face feature model and the out block. In model 1, the output block using the emotion category classification branch. About model 2, it used all branches (emotion category classification, arousal-valence CCC, and MSE regression). The model 2's output is the expression and model 1's output are both expression and valence-arousal value.

Similarity, model 3 (Emotion Frame) and 4 (Emotion VA Frame) have the STAT module. Model 5 (Emotion Frame with LSTM) and 6 (Emotion VA Frame with LSTM) have both STAT and LSTM modules. They received the current frame and previous block frames with size 16 or 32 and returned the basic expression (model 3, 5), both basic expression and valence-arousal values (model 4, 6).

\subsection{Evaluation Metrics}

For track 2 with basic emotion recognition, ABAW Challenge used the average of accuracy and F1 score metrics in all frame-by-frame evaluation in the testing set. After that, the expression score is calculated as below equation:

\begin{equation}
    Score_{expression} = 0.67 * F_{1} + 0.33 * Accuracy
    \label{eq:expr_score}
\end{equation}

For track 1 with valence-arousal regression, ABAW Challenge used CCC metrics as follows: 
\begin{equation}
    Score_{CCC} = \frac{2s_{xy}}{\sigma_{x}^{2} + \sigma_{y}^{2} + \left(\bar{x}-\bar{y}\right)^{2}}
    \label{eq:concordance_correlation_coefficient}
\end{equation}
where $s_{x}$ and $s_{y}$ are the variances of all frames in video with the valence/arousal annotations and predicted values, respectively, $\bar{x}$ and $\bar{y}$ are their corresponding mean values and $s_{xy}$
is the corresponding covariance value.

Finally, the mean value of CCC in valence and arousal will be used as main score in the track 1.
\begin{equation}
    Score_{CCC, total} = \frac{Score_{CCC,Arousal} + Score_{CCC, Valence}}{2}
    \label{eq:concordance_correlation_coefficient}
\end{equation}

\subsection{Results and Discussion}

\begin{table}[H]
\centering
\caption{Result Comparison in Expression Score, and Valence-Arousal Score on Validation set with baseline in \cite{Kollias2020}}
\label{tbl:list_results}
\begin{tabular}{|c|l|c|c|c|c|c|}
\hline
\textbf{Model}   & \textbf{Acc.} & \textbf{F1} & \textbf{Expr. Score} & \textbf{Aro.} & \textbf{Val.} & \textbf{VA Score} \\ \hline
\cite{Kollias2020}  & - & - & 0.36 & 0.14     & 0.24     & 0.19     \\ \hline
1   & 0.408 & 0.417 & 0.409 & -     & -     & -     \\ \hline
2   & 0.493 & 0.512 & \textbf{0.501} & 0.484 & 0.484 & 0.484 \\ \hline
3   & 0.401 & 0.414 & 0.405 & -     & -     & -     \\ \hline
4   & 0.479 & 0.507 & 0.492 & 0.564 & 0.504 & \textbf{0.534} \\ \hline
3   & 0.4 & 0.405 & 0.399 & -     & -     & -     \\ \hline
6   & 0.428 & 0.441 & 0.432 & 0.57 & 0.458 & 0.514 \\ \hline
Fusion   & 0.534 & 0.556 & 0.543 & 0.56 & 0.495 & 0.527 \\ \hline
\end{tabular}
\end{table}

We used the average fusion with the result for the expression score \textbf{0.543}, and valence-arousal score \textbf{0.527}. As Table \ref{tbl:list_results}, we achieved the high accuracy in model 2 for track 2 with 0.501, and model 4 for track 1 with 0.534.

\begin{table*}
\centering
\caption{Result Comparison in Expression Score, and Valence-Arousal Score on Test set with baseline in \cite{Kollias2020}}
\label{tbl:list_test_results}
\begin{tabular}{|c|c|c|l|c|c|c|c|c|}
\hline
\textbf{Track} & \textbf{Submission}   & \textbf{Model}   & \textbf{Acc.} & \textbf{F1} & \textbf{Expr. Score} & \textbf{Aro.} & \textbf{Val.} & \textbf{VA Score} \\ \hline
1 & - & \cite{Kollias2020}  & - & - & - & 0.27  & 0.11     & 0.19   \\ \hline
1 & 1 & 2                   & - & - & - & 0.295 & 0.356    & 0.325  \\ \hline
1 & 2 & 4                   & - & - & - & 0.342 & 0.368    & 0.355  \\ \hline
1 & 3 & 6                   & - & - & - & 0.383 & 0.381    & \textbf{0.382} \\ \hline
1 & 4 & Fusion              & - & - & - & 0.354 & 0.386    & 0.37   \\ \hline
2 & - & \cite{Kollias2020}  & - & - & 0.30 & -  & - & -  \\ \hline
2 & 1 & 2                   & 0.546 & 0.263 & 0.356 & -  & - & -  \\ \hline
2 & 2 & 4                   & 0.48  & 0.264 & 0.335 & -  & - & -  \\ \hline
2 & 3 & 2                   & 0.547 & 0.311 & \textbf{0.389} & -  & - & -  \\ \hline
2 & 4 & Fusion              & 0.565 & 0.295 & 0.384 & -  & - & -  \\ \hline
\end{tabular}
\end{table*}

In Table \ref{tbl:list_test_results}, we showed our results for the submission on Track 1 (Valence-Arousal Challenge) and Track 2 (Expression Challenge). For Track 1, we only used the models 2 (Emotion VA Image), 4 (Emotion VA Frame), 6 (Emotion VA Frame with LSTM). The best result \textbf{0.382} achieved with the model  6 on the test set. For Track 2, the model 2 achieved the best result \textbf{0.389}. 

\section{CONCLUSIONS AND FUTURE WORKS}
In this paper, we presented the effective method for affective behavior analysis in-the-wild on Aff-Wild2 dataset. It contains the temporal and stat module to exploit and fine-tune again the face feature extraction model. We achieved accuracy higher than the baseline model on track 1 and 2 of the ABAW Challenge.

\section{ACKNOWLEDGMENTS}

This research was supported by Basic Science Research Program through the National Research Foundation of Korea (NRF) funded by the Ministry of Education (NRF-2017R1A4A1015559, NRF-2018R1D1A3A03000947).








\bibliography{main}{}

\begin{thebibliography}{10}\itemsep=-1pt

\bibitem{cao2018vggface2}
Q.~Cao, L.~Shen, W.~Xie, O.~M. Parkhi, and A.~Zisserman.
\newblock {VGGFace2: A Dataset for Recognising Faces across Pose and Age}.
\newblock In {\em IEEE International Conference on Automatic Face {\&} Gesture
  Recognition (FG)}, pages 67--74. IEEE, IEEE, may 2018.

\bibitem{Dhall2019b}
A.~Dhall, S.~Ghosh, R.~Goecke, and T.~Gedeon.
\newblock {EmotiW 2019: Automatic emotion, engagement and cohesion prediction
  tasks}.
\newblock In {\em ICMI 2019 - Proceedings of the 2019 International Conference
  on Multimodal Interaction}, 2019.

\bibitem{dhall2012collecting}
A.~Dhall, R.~Goecke, S.~Lucey, T.~Gedeon, A.~Dhall, S.~Member, S.~Lucey,
  T.~Gedeon, R.~Goecke, S.~Lucey, T.~Gedeon, and Others.
\newblock {Collecting large, richly annotated facial-expression databases from
  movies}.
\newblock {\em IEEE Multimedia}, 19(3):34--41, jul 2012.

\bibitem{Ekman1992b}
P.~Ekman.
\newblock {An Argument for Basic Emotions}.
\newblock {\em Cognition and Emotion}, 1992.

\bibitem{He2016}
K.~He, X.~Zhang, S.~Ren, and J.~Sun.
\newblock {Deep Residual Learning for Image Recognition}.
\newblock In {\em IEEE conference on computer vision and pattern recognition
  (CVPR)}, pages 770--778, 2016.

\bibitem{kollias2017recognition}
D.~Kollias, M.~A. Nicolaou, I.~Kotsia, G.~Zhao, and S.~Zafeiriou.
\newblock {Recognition of Affect in the Wild Using Deep Neural Networks}.
\newblock In {\em IEEE Computer Society Conference on Computer Vision and
  Pattern Recognition Workshops}, volume 2017-July, pages 1972--1979, 2017.

\bibitem{Kollias2020}
D.~Kollias, A.~Schulc, E.~Hajiyev, and S.~Zafeiriou.
\newblock Analysing affective behavior in the first abaw 2020 competition.
\newblock {\em arXiv preprint arXiv:2001.11409}, 2020.

\bibitem{kollias2019deep}
D.~Kollias, P.~Tzirakis, M.~A. Nicolaou, A.~Papaioannou, G.~Zhao, B.~Schuller,
  I.~Kotsia, and S.~Zafeiriou.
\newblock {Deep Affect Prediction in-the-Wild: Aff-Wild Database and Challenge,
  Deep Architectures, and Beyond}.
\newblock {\em International Journal of Computer Vision}, 127(6-7):907--929,
  2019.

\bibitem{Kollias2018c}
D.~Kollias and S.~Zafeiriou.
\newblock Aff-wild2: Extending the aff-wild database for affect recognition.
\newblock {\em arXiv preprint arXiv:1811.07770}, 2018.

\bibitem{Kollias2018b}
D.~Kollias and S.~Zafeiriou.
\newblock A multi-task learning \& generation framework: Valence-arousal,
  action units \& primary expressions.
\newblock {\em arXiv preprint arXiv:1811.07771}, 2018.

\bibitem{Kollias2019a}
D.~Kollias and S.~Zafeiriou.
\newblock Expression, affect, action unit recognition: Aff-wild2, multi-task
  learning and arcface.
\newblock {\em arXiv preprint arXiv:1910.04855}, 2019.

\bibitem{Plutchik1984}
R.~Plutchik.
\newblock {Emotions: A General Psychoevolutionary Theory}.
\newblock In {\em Approaches to Emotion}. 1984.

\bibitem{zafeiriou2017aff}
S.~Zafeiriou, D.~Kollias, M.~A. Nicolaou, A.~Papaioannou, G.~Zhao, and
  I.~Kotsia.
\newblock {Aff-Wild: Valence and Arousal 'In-the-Wild' Challenge}.
\newblock In {\em IEEE Computer Society Conference on Computer Vision and
  Pattern Recognition Workshops}, volume 2017-July, pages 1980--1987. IEEE, jul
  2017.

\end{thebibliography}
\bibliographystyle{ieee}

\end{document}